\documentclass[runningheads]{llncs}
\usepackage{amsmath} 
\usepackage{amssymb} 
\usepackage{flexisym}
\usepackage[boxruled, linesnumbered]{algorithm2e}
\usepackage{multicol}
\usepackage{graphicx,subfigure} 
\usepackage{multirow}
\usepackage{booktabs}
\usepackage{enumitem}
\usepackage{tabularx}
\usepackage{url}

\def\gt{ground-truth}
\def\hl{Huber loss}
\def\od{object detection}
\def\bboxes{bounding boxes}
\def\bbox{bounding box}
\def\iou{IoU}
\def\fig{Fig.~}

\def\iouswitch{Smooth IoU loss}

\begin{document}
\title{Directly Optimizing IoU for Bounding Box Localization}
%
%
\author{Mofassir ul Islam Arif\inst{1}\and
Mohsan Jameel\inst{1} \and
Lars Schmidt-Thieme\inst{1}}
\authorrunning{Arif. Author et al.}
%
\institute{Information Systems and Machine Learning Lab (ISMLL)\\
 Univerity of Hildesheim, Hildsheim, Germany\\
\email{\{mofassir,mohsan.jameel,schmidt-thieme\}@ismll.uni-hildesheim.de}}
\maketitle              
\begin{abstract}
Object detection has seen remarkable progress in recent years with the introduction of Convolutional Neural Networks (CNN). Object detection is a multi-task learning problem where both the position of the objects in the images as well as their classes needs to be correctly identified. 
The idea here is to maximize the overlap between the \gt~bounding boxes and the predictions i.e. the Intersection over Union (IoU). 
In the scope of work seen currently in this domain, IoU is approximated by using the Huber loss as a proxy but this indirect method does not leverage the \iou~information and treats the \bbox~as four independent, unrelated terms of regression. This is not true for a \bbox~where the four coordinates are highly correlated and hold a semantic meaning when taken together. The direct optimization of the \iou~is not possible due to its non-convex and non-differentiable nature.
In this paper, we have formulated a novel loss namely, the Smooth IoU, which directly optimizes the IoUs for the bounding boxes. This loss has been evaluated on the Oxford IIIT Pets, Udacity self-driving car, PASCAL VOC, and VWFS Car Damage datasets and has shown performance gains over the standard \hl.

\end{abstract}
\section{Introduction}
\label{sec:intro}


Object detection is a multi-task learning problem with the goal of correctly
identifying the object in the image while also localizing the object into a
bounding box, therefore the end result of the object detection is to classify
and localize the object. As with all machine learning models, the optimization
is dictated by a loss that updates a loss towards a local optimum solution.
The family of object detection models \cite{Girshick_2015_ICCV} 
\cite{ren2015faster} \cite{liu2016ssd} \cite{redmon2017yolo9000} is accompanied by 
multi-task \cite{caruana1997multitask} losses which are made up of a
localization loss $\mathcal{L}_{loc}$ and a classification loss
$\mathcal{L}_{cls}$, for each stage. For the first stage the
$\mathcal{L}_{loc}$ is used to distinguish between the raw proposals from a
Region Proposal Network (RPN) usually modeled by a Fully Convolutional Network (FCN) \cite{long2015fully},
and the ground truth \bboxes. The aim here is to separate the background and
the foreground, based on the bounding boxes, therefore, the classification
loss $\mathcal{L}_ {cls}$ becomes a binary classification problem between the
foreground and the background. The output of this stage is passed to the
second stage where second stage localization and classification losses are
used. In the second stage, \bbox~localization deals with the actual
objects rather than the background and foreground. Similarly the second stage
classification loss is now a $K$-way softmax where $K$ is the number of
classes. For each stage, these losses are jointly optimized during training by
forming a linear combination of the two, therefore the total loss for each
stage is: 
\begin{equation} 
\mathcal{L} = \mathcal{L}_{loc} + \mathcal{L}_{cls}
\end{equation}
During training, \gt~\bboxes~are
used to train the model to learn the features of the objects that are present
within the constraints of the boxes. Traditionally the two-stage methods rely
on the \hl~\cite{box1956annals} for \bbox~localization in both stages. Eq. 
\ref{eq:huberloss} shows the \hl, its popularity  in R-CNN, Fast RCNN,
Faster-RCNN, SSD and many others is due to its robustness against outliers. In
our case, the outliers would be the \bboxes~that are very far away from the
\gt. 

\begin{equation} 
L_\delta (z)= \begin{cases} \frac{1}{2}z^2, & \text{if}\ |z|<\delta \\ 
\delta|z| - \frac{1}{2}\delta^2, & \text{otherwise} 
\end{cases}
\label{eq:huberloss}  
\end{equation}\break
\begin{equation}
BB_{regression} = \min_\theta \mathcal{L}( \beta ,\hat{\beta}(\theta))
\label{eq:bb_formal}
\end{equation}

Here $z$ is the L1 loss between the \gt~ and
predicted \bbox and $\delta$ is a threshold parameter. The \bbox~localization therefore, is treated
as a regression problem as seen in Eq. \ref{eq:bb_formal}.
Where $\beta$ is the ground truth \bbox~and  $\hat{\beta}(\theta)$ is the
prediction model, parametrized by $\theta$ which are the parameters learned
during the training phase, the output is predicted bounding boxes. Each
\bbox~is a tuple $((x_{1},y_{1}),(x_{2},y_{2}))$ which represents the
coordinates on the diagonal of the box. This regression problem deals with each
of the four
parameters of the \bbox~as independent and unrelated items however
semantically that is not the case since the four coordinates of the \bbox~are
highly correlated and need to be treated as a single entity. 


\begin{figure}
\centering     
\subfigure{\label{fig:toy}}\includegraphics
[width=0.49\linewidth]{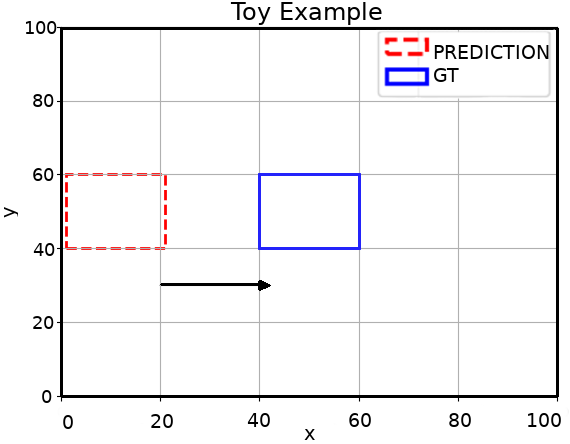}
\subfigure{\label{fig:loss}}\includegraphics[width=0.49\linewidth]
{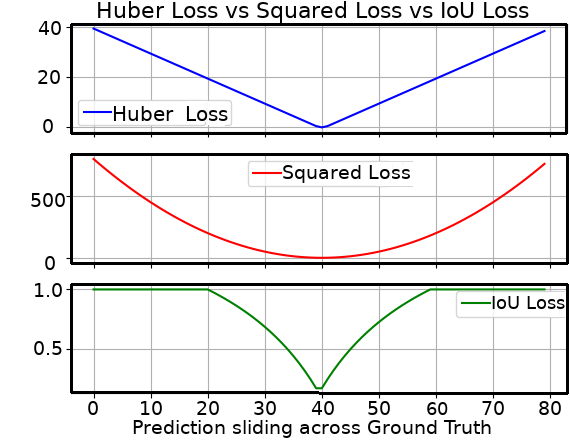}
\caption{The left figure mimicks the behaviour of a model predicting
incrementally correct bounding boxes. The prediction is 'slid' over the \gt~
to examine the effect on the different losses. The figure on the right shows
the behavior of the losses.}
\label{fig:losses}
\end{figure}

Huber loss, used in \bbox~localization, has a quadratic behavior for values
$|z| < \delta$ which
enables faster convergence when the difference (location, size, scale) between the
\gt~ and predictions
become small. For the regions where the difference between the boxes is
greater than the threshold $\delta$ the \hl~evaluates the L1 loss which has
been shown to be  less aggressive against outliers, this prevents exploding gradients due to large penalties, a behavior that is
seen by the penalty incurred by the squared loss.
While this loss has shown to be a good surrogate by casting the 
maximization of \iou~between \gt~and predicted \bboxes~as a four-point
regression, it does not use the IoU
information during optimization. Furthermore, as stated earlier, it conducts
the \bbox~regression without considering the parameters of a \bbox~to be highly
correlated items which hold a semantic meaning when taken together. 
Therefore, it stands to reason that the optimization of the \od~loss, more
specifically the \bbox~localization should involve a direct
optimization of the \iou. The calculation  of the \iou~can be seen
in Eq. \ref{eq:iou_loss}. Here $\beta = ((x_1,y_1),(x_2,y_2))$ is the \gt~\bbox~and $
\hat{\beta} = ((\hat{x}_1,\hat{y}_1),(\hat{x}_2,\hat{y}_2))$ is the predicted \bbox.
\begin{equation}
\begin{gathered}
{IoU} = \frac{I_w \times I_h }{Area_1 + \hat {Area}_2 -
(I_w \times I_h)}
\label{eq:iou_loss}
\end{gathered}
\end{equation}
The areas for the \bboxes~are calculated as $Area= (x_2 - x_1)\times
(y_2 - y_1)$.
Converting the \iou~measure into a loss function is trivial, since 
$\mathcal{L}_{\iou}= 1 -\iou$. 
The intersection term is calculated based on the region of overlap between the
two boxes and it is as follows:
\begin{equation}
\begin{gathered}
{Intersection} = I_w \times I_h
\label{eq:iou_intersection}
\end{gathered}
\end{equation}

Here, $I_h = max(0, min_y - max_y)$ is the intersection height and  where $min_y =  min(y_2,\hat y_2)$ and $max_y =
 max(y_1,\hat y_1)$ are the minimum and maximum y-coordinates, respectively. Similarly, $I_w =
max(0, min_x - max_x) $ is the intersection width with $min_x = 
 min(x_2,\hat x_2)$ and $max_x =  max(x_1,\hat x_1)$ as the minimum and maximum
x-coordinates for the overlapping region, respectively. The product of $I_h$ and $I_w$
as in Eq. \ref{eq:iou_intersection}, yeilds the intersection. The denominator term for Eq.
\ref{eq:iou_loss} shows the union term, here $Area_1$ and $Area_2$ are the \gt~and predicted \bbox~areas, respectively.


An examination of the behaviors for the different losses can be seen in \fig \ref{fig:losses}. The example presented is
designed to show the behavior of the losses as a model predicts \bboxes~that are translated over the \gt~\bbox. The two
boxes are not overlapping up until the point that the predicted \bbox~reaches $(20,40)$. At this point, the overlap
starts to occur and increases until the point $(40,40)$, at which the overlap is maximum and starts to decrease as the
box continues moving to the right. The overlap drops to zero at point $(60,40)$. For the sake of simplicity of the illustration, we are
limiting the movement of the box to the x coordinate only in Fig. \ref{fig:losses}. 


For \hl, we can see a linear decrease as the predicted box approaches the
\gt~and conversely the loss linearly increases as the box starts to exit from
the \gt. While inside some threshold $\delta$ it behaves quadratically, for
areas with no overlap we can see the robustness of the \hl~as it does not incur
a large loss value. The L2 loss shows a similar (increase and decrease) behavior
but is not bound by any $\delta$ parameter and is, therefore, quadratic
throughout. This leads to a very high penalty around the area where the boxes
do not overlap. These effects are obvious when we compare the scales of the
losses.

Lastly, we can see the \iou~loss plateauing outside the region where the two \bboxes~do not overlap. Here it can be seen
that for the regions with no overlap between the two boxes the loss plateaus leading to zero gradients thus, effectively
making the learning process impossible for all gradient-based learning methods. The areas outside the intersection offer
no help in the learning process due to the fact that the \iou~is bounded in $[0,1]$ and the worst case scenario i.e, no
overlap always leads to a loss of 1, regardless of how far away the box is. From the loss profile, we can also see that
the \iou~loss is non-convex (due to the plateauing region violating Eq. \ref{eq:convex}) and non-differentiable.
\begin{equation}
f(tx_1 + (1-t)x_2) \leq tf(x_1) + (1-t)f(x_2)
\label{eq:convex}
\end{equation}
The \hl~does not suffer from these issues
since the regression always returns the distance between the parameters of the \bboxes.
The shortcoming here is the inherent treatment of the
parameters of the \bboxes~as independent and unrelated terms. 
 
In this paper, we present a novel loss that addresses the shortcoming of the standard IoU loss, inherits the advantages
of \hl,~and enables a direct optimization of \iou~for two-stage \od~networks. This is done by a proposed  relaxation for
the \iou~loss which mitigates the non-differentiability and non-convexity of the loss without the need to sub-gradient
or approximation methods \cite{nedic2001incremental}. We propose a dynamic loss, that leverages the gains of \hl~while 
directly optimizing for \iou~in \bbox~localization. The main contributions of this
paper are:

\begin{enumerate}[label=\alph*)]  \item A robust loss that can be integrated readily into the two-stage models. \item A
performance guarantee that is lower bounded by the state-of-the-art performance. \item  Empirical analysis of the
proposed method on standard \od~datasets to show how optimizing for \iou~can lead to better bounding boxes (higher IoU).
\end{enumerate}

\section{Related Work}


The choice of losses in machine learning is dictated heavily by the convergence and convexity of the loss
\cite{rosasco2004loss}. The \hl~ensures a stable convergence due to its piece-wise quadratic and convex  nature
\cite{pmlr-v70-xu17a}. This enabled the loss to be adapted readily in the bounding box regression for the object
detection tasks \cite{Girshick_2015_ICCV} \cite{ren2015faster} \cite{liu2016ssd}. Similar to the \hl~ there is also an
interest in using the squared loss for the bounding box regression  \cite{erhan2014scalable} \cite{szegedy2014scalable}.
This is however more susceptible to exploding gradients and is more sensitive to the learning rate and other such
hyper-parameters \cite{schaul2013no}. In \fig\ref{fig:losses} we have demonstrated the behavior of these losses and how
they, while suitable for regression, optimize for a proxy loss and a more direct approach for optimizing the IoU is
needed. The disadvantage of the \iou~loss  stems from its non-differentiability and the disappearing gradients outside
the regions of intersection.  Attempts to addresses this problem by using the IoU loss by looking only at the pixel
values inside the predictions and ground truth boxes with a non zero overlap \cite{DBLP:journals/corr/YuJWCH16}. They
convert the IoU loss into a cross-entropy loss since $0 \leq IoU \leq 1$ by wrapping it in the natural log
$\mathcal{L}=-ln(IoU)$. This conversion relies on using the IoU information after converting the tuple of four
coordinates into a pixel map and then evaluating the IoU pixel-wise. Furthermore, they propose a novel architecture for
their loss implementation thus might not be readily compatible with the other architectures used for the object
detection tasks. Another related method looks at the complete replacement of the regression loss with their
implementation of the IoU loss \cite{rahman2016optimizing}. They approach the task of image segmentation and how the
optimization of the IoU directly can serve to improve the overall performance of the model. They rely on a FCN which is
a modified AlexNet \cite{sutskever2012imagenet} and present the work in light of how for the image segmentation task,
the discrimination between the background and the foreground serves as an important step.  However, optimizing for the
overall accuracies could cause a model to encourage larger sized boxes. This can be the case when a larger portion 
($90\%$) of the image belongs to the background, in such a situation a naive algorithm can get $90\%$ accuracy simply by
predicting everything to be the background \cite{DBLP:journals/corr/YuJWCH16}. A case like this can be made for using
the \hl~for the bounding box regression which the \hl~treats four independent and unrelated items during its
optimization. The use of Bayesian decision theory has also been attempted by  \cite{nowozin2014optimal} where
Conditional Random Field (CRF) is used to maximize the expected-IoU, they also use the pixel values and a greedy
heuristic for the optimization of IoU. A pixel-wise approach is inherently slower since it is dictated by the number of
pixels in the \bbox. A \bbox with size $P \times Q$ where $P$ is the width and $Q$ is the heigth would require $O(PQ)$
operations in order to calculate the \iou. Whereas, by treating the \bbox~as a tuple and calculating \iou~ as in Eq.
\ref{eq:iou_loss} the number of operations is constant regardless of the size.

\section{Methodology}


The \iou~loss suffers from the plateauing phenomenon because of the
unavailability of gradient information since \text{$L_{\iou} \in [0,1]$} where
it is constant (1) outside the region of intersection as shown in
\fig\ref{fig:losses}. This gradient information in a standard \hl,~for
\bbox~localization, is available throughout due to the regression between the
four points of the \gt~and predicted \bbox. The vanishing gradients of
\iou~loss for \bboxes~with no overlap hinder the learning process since two
\bboxes~with no overlap present the same constant loss (zero gradients) regardless of how far they are
from the \gt.
A relaxation is needed for the \iou~loss that will
enable us to bring in the gradient information for the predicted bounding box
in terms of the distance, and consequently guide the model in the correct
direction. Albeit, this is needed only in the initial learning stage because
once the predicted bounding boxes begin to overlap with the ground truth ones, non-zero overlap will address the
plateauing behavior of the \iou~loss. 
Standard object detection models
treat the regression as an
independent and unrelated four-way entity which is not true for a bounding
box.

In order to optimize the true goal of object detection, we need to directly
optimize for \iou~in the \bbox~localization loss. Our method proposes to morph the \hl~in order
to include the \iou~information.

\subsection{Smooth IoU Loss}


A smooth stiched loss, named Smooth IoU is presented as an improvement on the
\hl~to enhances the localization of the \bboxes~while also overcoming the
non-convexity issues (stemming from the \iou~loss being bounded in $[0,1]$) of
the vanilla \iou~loss, and is presented in Eq. \ref{eq:smoothiou}.

\begin{equation}
\mathcal{L}_{SmoothIoU} = \lambda \mathcal{L}_{IoU} + (1 - \lambda)\mathcal{L}_{HuberLoss}
\label{eq:smoothiou}
\end{equation}

The first term of Eq. \ref{eq:smoothiou} is the \iou~element which directly
incorporates the \iou~in the optimization process. The second term is the
state-of-the-art \hl. The purpose of having the \hl~is to make sure that the
positional guidance can be made use when there is no overlap between the
\gt~and predicted \bboxes, thus making gradient information available
throughout the learning process. 
The two terms of the loss are linked by a scaling parameter $\lambda$.
Naively, this term can be treated as a hyper-parameter that can be tuned for
the best performance, however, doing so will be computationally expensive as
well as time-consuming.
Additionally, treating $\lambda$ as a hyper-parameter will lead to
having one $\lambda$ for the entire retraining which was found to be
detrimental to the overall performance. A mini-batch could have poor
predictions and thus lead to \bboxes~with no overlap in which case the fixed
value for $\lambda$ would still try to make use of the non-existent gradients
coming from the \iou~element of Eq. \ref{eq:smoothiou}. In order to prevent
such outcomes, we propose to treat $\lambda$ not as a hyper-parameter but
rather scale it dynamically during training. $\lambda$ is calculated based on
the mean \iou~of the minibatch under evaluation and used in scaling the loss
between \iou~and state-of-the-art \hl.
\begin{figure}[t]
\centering     
\subfigure{\label{fig:toy}}\includegraphics
[width=0.49\linewidth]
{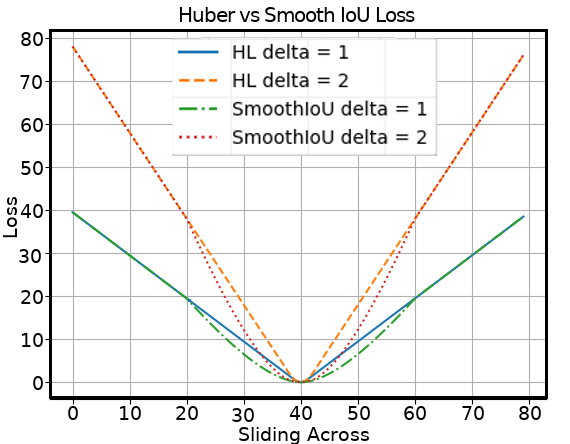}
\subfigure{\label{fig:loss}}\includegraphics[width=0.49\linewidth]
{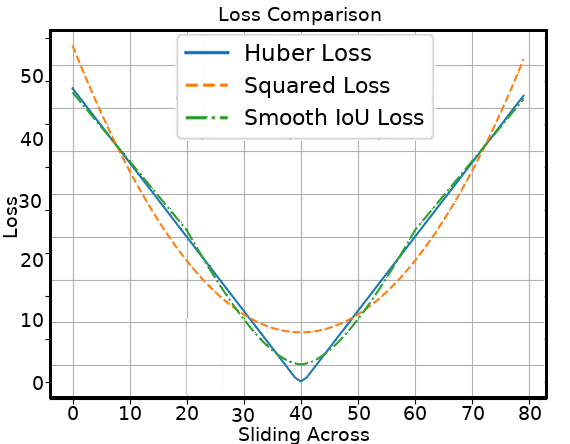}
\caption{(Right) Comparison of the Smooth IoU with Huber and Squared
Loss, the losses have been scaled to highlight the
profiles as they relates to the others. (Left) Behavior of Huber
and Smooth IoU losses for varying
values of $\delta$. (see Eq. \ref{eq:huberloss}).}
\label{fig:smooth_smooth}
\end{figure}
This dynamic scaling
enables us to remove the need to tune $\lambda$ and allows the model to learn
End-to-End. This enables the model to be trained faster and without the need
for a problem specific set of hyper-parameters.

The loss profile for the \iouswitch~is presented in
\fig\ref{fig:smooth_smooth} (left). In order to distinguish our contribution
from that state-of-the-art \hl, we are presenting the behavior of the losses
for the same example as seen in \fig \ref{fig:losses} where a predicted
\bbox~is translated over the ground truth \bbox. The $\delta$ term is the
cutoff threshold from Eq. \ref{eq:huberloss} and is varied from 1-2 in order
to show that the behavior of the \iouswitch~is not just a scaled variation of
the inherent \hl~cutoff behavior. \fig \ref{fig:smooth_smooth} (left) shows the
behavior of the \iouswitch~and highlights how it is purely a function of the
overlap of the boxes. From the figure, it can be seen that the
\iouswitch~introduces a quadratic behavior to the loss as soon as the overlap
starts to occur at the point $(20,40)$ (see the description of the example in
the introduction). This quadratic behavior appears much later in the
state-of-the-art \hl, which is governed by the $\delta$ and is not dynamic but
rather fixed for each run. Outside of the areas
of intersection, we are still maintaining the \hl~which allows the loss to
overcome the inherent shortcomings of the \iou~loss, and localize the boxes
better.

\begin{algorithm}

 \KwData{predicted\_boxes $P := \{p \in \mathbb{R}^{K\times4} \ |\  p =  \{y_1,x_1,h,w\}\}$ ,  target\_boxes $T := \{t \in \mathbb{R}^{K\times4} \ |\  t = \{\bar y_1,\bar x_1,\bar h, \bar w\}\} $} \KwResult{Smooth IoU Loss}

\For{k = 1, \dots ,K}{
P\textprime = Transform ($P_k$)\;  
T\textprime = Transform ($T_k$)\;
$\mathcal{L}_{Huber_k}$ = Huberloss ($P_k\textprime,T_k\textprime$)\;
$IoU_k$ = Calculate\_IoU($P_k\textprime,T_k\textprime$)\;
$\mathcal{L}_{IoU_k}$ = 1 - $IoU_k$\;
}

$\lambda$ = mean($IoU_{1:K}$)\;
\For{k = 1, \dots, K}{
$loss_{SmoothIoU_k}$ = $\lambda \times \mathcal{L}{IoU_k}$ + (1- $\lambda$)$\mathcal{L}_{Huber_k}$\;
}
\caption{Smooth IoU Loss}
\label{algo:smooth}
\end{algorithm}

The \iouswitch~comes with a performance guarantee of the state-of-the-art in
the worst case scenario i-e. the loss converges to the state-of-the-art
performance if the modifications suggested in this method do not improve the
\bbox~localization. This can be shown as:

\begin{equation}
\lim_{\lambda\to0} \mathcal{L}_{SmoothIoU} = \mathcal{L}_{HuberLoss}
\end{equation}
\begin{equation}
\lim_{\lambda\to 1} \mathcal{L}_{SmoothIoU} = \mathcal{L}_{IoU}
\end{equation}

This further highlights that the loss presented here is guaranteed to perform
at the state-of-the-art level in the worst case scenario, making it a robust
version of the \hl~while introducing the \iou~information into the optimization
process. Additionally, this loss can be readily substituted into the current
two-stage models without any architectural changes or case-specific
modifications, making it modular.

Algorithm \ref{algo:smooth}. shows the implemenetation of the Smooth IoU loss, lines 2-3 are used to transform the incoming points from $\{ y_1, x_1, h, w\}$ representation to a $\{ x_{min}, y_{min}, x_{max}, y_{max}\}$ representation. This is done in order to calculate the IoU of the bounding boxes. Line 10 shows the implementation of Eq. \ref{eq:smoothiou}.

\begin{figure}[b!]
\centering     
\subfigure{\label{fig:smoothiouv2}}\includegraphics
[width=0.49\linewidth]
{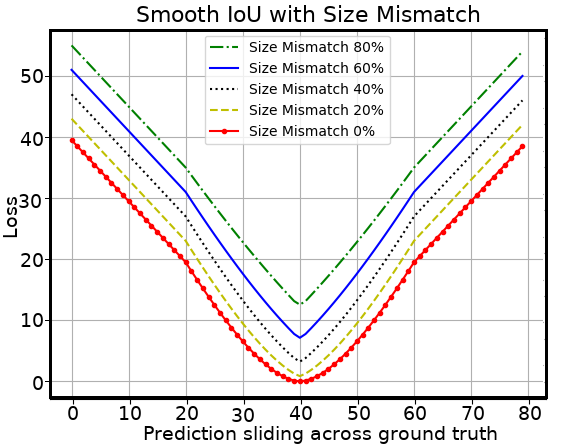}
\subfigure{\label{fig:loss}}\includegraphics[width=0.49\linewidth]
{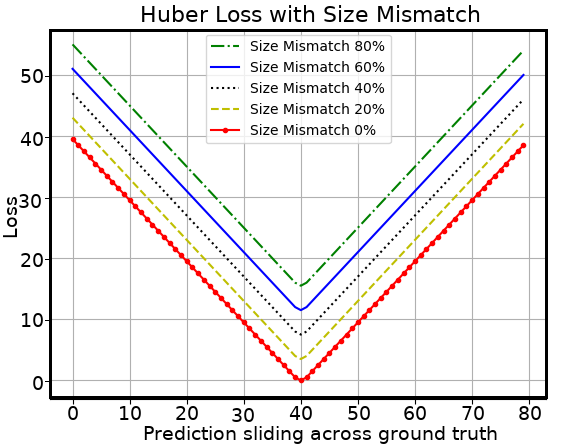}
\caption{Comparison of the losses with a size mismatch between the predicted
and ground truth \bboxes.}
\label{fig:l1mismatch}
\end{figure}

For the sake of completeness, the same example as seen in \fig\ref{fig:losses} is
reproduced but this time with an introduced size mismatch between the prediction
and ground truth \bboxes. The loss profiles for both the \iouswitch~and \hl~are
shown in \fig\ref{fig:l1mismatch}. The overall effect visible here is that even
for the size mismatch between the two boxes the \iouswitch~tends to converge to
a smaller loss value while leveraging the advantages of the \hl~for non-overlapping regions. 
This smaller loss value will prevent the possibility
of exploding gradients thus stabilizing the learning process.

\section{Experiments}


Having laid out the design of our new loss and its characteristics. We evaluated
the performance on the object detection task using the Oxford-IIIT Pet Dataset
\cite{parkhi12a}, Udacity Self-Driving Car Dataset \cite{udacity}, PASCAL VOC 
\cite{pascal-voc-2012}, and Volkswagon Financial Services (VWFS) Damage
Assessment dataset (propriety). 
\begin{table}[t!]
\footnotesize
\centering
\begin{tabular}{ll|l|l|l|}
\cline{3-5}
    &  & \textbf{\begin{tabular}[c]{@{}l@{}}Initial LR \end{tabular}} & \textbf{Proposals} & \textbf{\begin{tabular}[c]{@{}l@{}}Drop-out\end{tabular}} \\ \hline
\multicolumn{1}{|l|}{\multirow{2}{*}{\textbf{Pets}}}    & Smooth & 0.0002  & 300     & {[}0.2, ... , 0.8{]}  \\ \cline{2-5} 
\multicolumn{1}{|l|}{}    & Huber  & 0.0002  & 300     & {[}0.2, ... , 0.8{]}  \\ \hline
\multicolumn{1}{|l|}{\multirow{2}{*}{\textbf{Udacity}}} & Smooth & 0.002   & 300     & {[}0.2, ... , 0.8{]}  \\ \cline{2-5} 
\multicolumn{1}{|l|}{}    & Huber  & 0.002   & 300     & {[}0.2, ... , 0.8{]}  \\ \hline
\multicolumn{1}{|l|}{\multirow{2}{*}{\textbf{VWFS}}}    & Smooth & 0.0002  & 300     & {[}0.2, ... , 0.8{]}  \\ \cline{2-5} 
\multicolumn{1}{|l|}{}    & Huber  & 0.0002  & 300     & {[}0.2, ... , 0.8{]}  \\ \hline
\multicolumn{1}{|l|}{\multirow{2}{*}{\textbf{PASCAL}}}  & Smooth & 0.0002  & 300     & {[}0.2, ... , 0.8{]}  \\ \cline{2-5} 
\multicolumn{1}{|l|}{}    & Huber  & 0.0002  & 300     & {[}0.2, ... , 0.8{]}  \\ \hline
\end{tabular}
\caption{Hyper-parameters used for the different datasets}
\label{tab:hyp}
\end{table}
The VWFS dataset is made up of images taken during an end-of-leasing damage
assessment. The areas of interest in these images are the damaged parts that
have been loosely annotated. The aim of this data is to serve as a foundation
for training a model which is able to detect these damages automatically and
estimate the damage costs at the end of the car lease period. The data has a
very high variance in the number of images per class and suffers and from a
long tail distribution. All experiments were conducted using Nvidia GTX
1080Ti, and Tesla P100 GPUs.

The experiments conducted herewith focused on Faster-RCNN from Tensorflow
Object Detection API \cite{huang2017speed} but \iouswitch~can be used with
any two-stage model. The reported results
are with the hyper-parameters reported in Tab. \ref{tab:hyp} tuned
to the best performance for the respective loss and to reproduce the numbers
reported in the original paper \cite{ren2015faster}. For the minimization of the loss
we have used RMSProp \cite{tieleman2014rmsprop}, with a learning rate reduced
by $10^{-1}$ every 50K steps and a momentum term of 0.9.
The underlying feature extractor was Inception_V2 \cite{ioffe2015batch},
pre-trained on the COCO dataset \cite{lin2014microsoft}. This warm start
enabled us to speed up training by leveraging the advantages offered by
transfer learning and as it has been shown to be an effective method for
initializing the network \cite{Girshick_2015_ICCV} \cite{ren2015faster}
\cite{huang2017speed}.  The baseline implementation \cite{ren2015faster} used
VGG-16 \cite{simonyan2014very} pre-trained on
ImageNet \cite{russakovsky2015imagenet} as the feature extractor.
In all of our experiments, the models were
retrained for 200k iterations and showed a smooth convergence behavior.

For the Udacity, and Pets datasets we have used a standard train/test split.
For the VWFS dataset, a 80-20 split was created while for the PASCAL VOC
dataset we have used the VOC2007 Train split for training the models and its
performance is presented on the VOC2007 val split. VOC2012 was treated the
same way. We have also used a merged PASCAL VOC dataset where the VOC++Train
was created by merging the VOC2007 and VOC2012 train splits for training and
we are presenting the results on the VOC2012 val dataset.  Since the ground
truth bounding box information was not available on the test set of PASCAL VOC
datasets, we present the results on the validation set.

\subsection{Evaluation and Discussion}

\begin{table}[t]
\begin{tabular}{|c|l|l|l|l|l|l|l|l|l|l|l|l|}
\hline
\multicolumn{1}{|l|}{} & \multicolumn{2}{c|}{\textbf{Pets}} & \multicolumn{2}
{c|}{\textbf{UD}} & \multicolumn{2}{c|}{\textbf{VW}} & \multicolumn{2}{c|}{
\textbf{2007}} & \multicolumn{2}{c|}{\textbf{2012}} & \multicolumn{2}{c|}{\textbf{VOC++}} \\ \hline
\textbf{$\mathcal{L}$}& \textbf{$\mathcal{L}_{HL}$}  & \textbf{$\mathcal{L}_{S\_{\iou}}$}  &\textbf{$\mathcal{L}_{HL}$} & \textbf{$\mathcal{L}_{S\_{\iou}}$}  & \textbf{$\mathcal{L}_{HL}$}   & \textbf{$\mathcal{L}_{S\_{\iou}}$}  & \textbf{$\mathcal{L}_{HL}$}  & \textbf{$\mathcal{L}_{S\_{\iou}}$}  & \textbf{$\mathcal{L}_{HL}$}  & \textbf{$\mathcal{L}_{S\_{\iou}}$}  & \textbf{$\mathcal{L}_{HL}$}  & \textbf{$\mathcal{L}_{S\_{\iou}}$}   \\ \hline
\textbf{IoU}  & 0.143  & \textbf{0.162}  & 0.417   & \textbf{0.425}      & 0.383
& \textbf{0.388}  & 0.231   & \textbf{0.263}  & \textbf{0.275}   & 0.270  &
0.240   & \textbf{0.244}   \\ \hline

\end{tabular} \caption{Localization Metrics: These highlight the results
for the datasets under test. \textbf{VW} = VWFS, \textbf{UD} = Udacity, \textbf{2007} = VOC2007, \textbf{2012} = VOC2012,
\textbf{VOC++} = VOC++Train. The IoU metric directly is reported here to show
the quality of the \bboxes. \textbf{$\mathcal{L}_{S\_ {\iou}}$} is optimized
for the IoU directly. \textbf{$\mathcal{L}_{HL}$} is the baseline \hl.}
\label{tab:loc_loss}  
\end{table} 
We propose a new loss for the \bbox~localization that takes into account the
direct optimization of the \iou~in order to improve the quality of the
predicted \bboxes. This localization loss ties in closely with the overall
performance of the two-stage networks. Therefore, it is important to evaluate
the \iou~quality of the model as well as the accuracies in order to showcase
the effectiveness of our method. Detection accuracies (mean average precision
and average recall) measure the correct classification of the objects and do
not directly take into account the quality of the predicted bounding boxes.
This is because mAP scores are calculated for a subset of the predicted
\bboxes~(that fall above a threshold of \iou).  We want to demonstrate how the
optimization of the \iou~directly for \bbox~localization helps in the overall
learning process while simultaneously improving the accuracy over the
state-of-the-art by proposing better \bboxes. Furthermore, we would also like
to showcase the robustness of the loss over varying levels of difficulty of the
object detection tasks, hence the choice of datasets that range from low
difficult (Pets) to high difficulty (PASCAL). As stated earlier, this loss
comes with a performance guarantee of the state-of-the-art performance and
that is shown in the results that follow. Faster RCNN with the standard
\hl~(for \bbox~localization) becomes the baseline.


We break down the evaluation of the model into localization and classification
performance. For the classification performance, we are using COCO detection
metrics as they are readily available in the API and are also a favored metric
for the object detection task. Localization is the primary focus here since
we are optimizing it directly for the \iou. Tab. \ref{tab:loc_loss} presents
the comparison of our proposed method against \hl~for the different datasets.
We are reporting the value of $\iou \in [0,1]$, the values in bold show where
our method is better than the baseline. We are evaluating the quality of the
\iou~against the \gt~\bboxes~(higher value is better). The reported numbers
here are all proposed boxes by the model, we are not discounting any boxes
through post-processing, hence the values appear to be small. This is done to
see the raw behavior of the model for the baseline and the proposed method. For
the localization, it can be seen in Tab. \ref{tab:loc_loss} that our method
outperforms the baseline in five out of six datasets that are under
consideration. The results show that by optimizing for the \iou~directly leads
to better \bboxes. Furthermore, the robustness of the loss is also verified by
looking at the results for VOC2012 in Tab. \ref{tab:cls_loss}. We underperform the baseline on the
VOC2012 dataset in terms of the \iou~however when we look at the overall
performance for the classification Tab. \ref{tab:cls_loss}, we can see that
for the VOC2012 dataset our method is still better than the baseline. This
indicates that the \iouswitch~can be used for directly optimizing the \iou~and
will not harm the overall performance in cases where it does not directly
improve the \iou.

\begin{table}[t]
\footnotesize
\centering
\begin{tabular}{|cl|lllll|}
\hline
Dataset  &   & \multicolumn{1}{c|}{\begin{tabular}[c]{@
{}c@
{}}mAP\\ @.50IoU*\end{tabular}} & \multicolumn{1}{c}{\begin{tabular}[c]{@{}c@{}}mAP\\ @.75IoU**\end{tabular}} & \multicolumn{1}{c}{mAP***} & \multicolumn{1}{c}{AR@1} & \multicolumn{1}{c|}{AR@10} \\ \hline

\hline
\hline
\multicolumn{1}{|l|}{\multirow{2}{*}{\textbf{Wins}}} & \textbf{$\mathcal{L}_{HL}$} & 0          & 3 & 2          & 2          & 1         \\ \cline{2-2} 
\multicolumn{1}{|l|}{} & \textbf{$\mathcal{L}_{S\_{\iou}}$}  & \textbf{6} & 3 & \textbf{4} & \textbf{4} & \textbf{5}  \\ \cline{2-2} 
\hline
\hline

\multicolumn{1}{|c|}{\multirow{2}{*}{\textbf{Pets}}} & \textbf{$\mathcal{L}_{HL}$}  & 89.94 & 80.46 & 66.03  & 75.59  & 76.63  \\ \cline{2-2} 
\multicolumn{1}{|c|}{} & \textbf{$\mathcal{L}_{S\_{\iou}}$}  & \textbf{93.91}   & \textbf{87.19}  & \textbf{72.09}  & \textbf{79.68}   & \textbf{80.35} \\ \cline{2-2} 
\hline
\multicolumn{1}{|c|}{\multirow{2}{*}{\textbf{VW}}}  & \textbf{$\mathcal{L}_{HL}$} & 64.3  & 37.01 & 37.03  & \textbf{21.0}   & 45.42  \\  
\multicolumn{1}{|c|}{} & \textbf{$\mathcal{L}_{S\_{\iou}}$}  & \textbf{64.79}  & \textbf{37.33}  & \textbf{37.4}   & 20.94  & \textbf{45.63}  \\ \cline{2-2} 
\hline
\multicolumn{1}{|c|}{\multirow{2}{*}{\textbf{UD}}}  & \textbf{$\mathcal{L}_{HL}$}  & 78.23  & \textbf{27.67}   & \textbf{36.22}  & 37.14  & 50.68  \\ \cline{2-2} 
\multicolumn{1}{|c|}{} & \textbf{$\mathcal{L}_{S\_{\iou}}$}   & \textbf{78.36,}  & 27.65   & 35.7  & \textbf{37.15} & \textbf{51.13} \\ \cline{2-2} 
\hline
\multicolumn{1}{|l|}{\multirow{2}{*}{\textbf{2007}}} & \textbf{$\mathcal{L}_{HL}$}  & 65.94  & \textbf{43.40}  & \textbf{40.68}  & \textbf{37.77}   & \textbf{57.14} \\ \cline{2-2} 
\multicolumn{1}{|l|}{} & \textbf{$\mathcal{L}_{S\_{\iou}}$}   & \textbf{66.29}   & 43.01  & 40.67  & 37.32   & 56.79 \\ \cline{2-2} 
\hline
\multicolumn{1}{|l|}{\multirow{2}{*}{\textbf{2012}}} & \textbf{$\mathcal{L}_{HL}$}   & 69.14  & \textbf{49.09}  & 44.20  & 39.63   & 59.82  \\ \cline{2-2} 
\multicolumn{1}{|l|}{} & \textbf{$\mathcal{L}_{S\_{\iou}}$}  & \textbf{69.31}   & 48.29  & \textbf{44.41}  & \textbf{40.26}   & \textbf{60.04} \\ \cline{2-2} 
\hline
\multicolumn{1}{|l|}{\multirow{2}{*}{\textbf{VOC++}}} & \textbf{$\mathcal{L}_{HL}$}  &70.47  & 49.68  & 44.19  & 39.81   & 59.39 \\ \cline{2-2} 
\multicolumn{1}{|l|}{} & \textbf{$\mathcal{L}_{S\_{\iou}}$}   & \textbf{70.52}   & \textbf{50.08}  & \textbf{45.17}  & \textbf{39.96}   & \textbf{59.58} \\ \cline{2-2} 

\hline
\end{tabular}
\caption{Classification Metrics:  *$mAP
@.50IoU$ is the PASCAL metric as set out in COCO Detections metrics, takes
into account \bboxes with a 50\% overlap with the
\gt. Similarly, **$mAP@.75IoU$ takes into account 75\% overlap.
 ***mAP takes into account overlap of 50\%
and higher. Average Recall (@1 and @10). }
\label{tab:cls_loss}
\end{table}
For the classification metrics in Tab. \ref{tab:cls_loss}, the first row
provides the total win/loss count for our proposed method and the baseline.
As can be seen, our method outperforms the baseline for mAP@.50IOU, mAP, AR@1,
and AR@10. We tie with the baseline for mAP@.75IOU. It should be noted here that
we are not modifying the baseline classification optimization and show how
directly optimizing the \iou~can lead to gains in the mAP and recall as well.
This lends credence to our proposed method shows it's usefulness. All the results highlight that there is a signal available in the IoU information of the bounding boxes and this information should be used during training. We experimentally showcase that the IoU driven loss variant proposed herewith can outperform the standard loss and it is lower-bounded to be at the state of the art level.

\section{Conclusion}


In this paper, we have presented a novel loss for the \bbox~localization of
two-stage models. The loss optimizes the \iou~directly by treating the
parameters of a \bbox~as a single highly correlated item. Our loss is
lower-bounded to perform at the state-of-the-art level. We demonstrate the
efficacy of our model by replacing the \hl~in Faster RCNN to show that
optimizing for \iou~directly in \bbox~localization can lead to better bounding
boxes and also improve the classification accuracy. Our method has shown to
outperform the baseline in both localization and classification metrics. The
modular and robust nature of the proposed loss makes it readily compatible
with all two-stage models. 

\section*{Acknowledgments}
This work is co-funded by the industry project "Data-driven Mobility Services" of ISMLL and Volkswagen Financial Services.(\url{https://www.ismll.uni-hildesheim.de/projekte/dna_en.html})
%
%
%
%
\bibliographystyle{splncs04}
\bibliography{egbib}

\begin{thebibliography}{10}
\providecommand{\url}[1]{\texttt{#1}}
\providecommand{\urlprefix}{URL }
\providecommand{\doi}[1]{https://doi.org/#1}

\bibitem{box1956annals}
Box, G., Hunter, J.: Annals of mathematical statistics. The Annals of
  Mathematical Statistics  \textbf{27}(4),  1144--1151 (1956)

\bibitem{caruana1997multitask}
Caruana, R.: Multitask learning. Machine learning  \textbf{28}(1),  41--75
  (1997)

\bibitem{erhan2014scalable}
Erhan, D., Szegedy, C., Toshev, A., Anguelov, D.: Scalable object detection
  using deep neural networks. In: Proceedings of the IEEE conference on
  computer vision and pattern recognition. pp. 2147--2154 (2014)

\bibitem{pascal-voc-2012}
Everingham, M., Van~Gool, L., Williams, C.K.I., Winn, J., Zisserman, A.: The
  {PASCAL} {V}isual {O}bject {C}lasses {C}hallenge 2012 {(VOC2012)} {R}esults.
  http://www.pascal-network.org/challenges/VOC/voc2012/workshop/index.html

\bibitem{Girshick_2015_ICCV}
Girshick, R.: Fast r-cnn. In: The IEEE International Conference on Computer
  Vision (ICCV) (December 2015)

\bibitem{huang2017speed}
Huang, J., Rathod, V., Sun, C., Zhu, M., Korattikara, A., Fathi, A., Fischer,
  I., Wojna, Z., Song, Y., Guadarrama, S., et~al.: Speed/accuracy trade-offs
  for modern convolutional object detectors. In: Proceedings of the IEEE
  conference on computer vision and pattern recognition. pp. 7310--7311 (2017)

\bibitem{ioffe2015batch}
Ioffe, S., Szegedy, C.: Batch normalization: Accelerating deep network training
  by reducing internal covariate shift. arXiv preprint arXiv:1502.03167  (2015)

\bibitem{lin2014microsoft}
Lin, T.Y., Maire, M., Belongie, S., Hays, J., Perona, P., Ramanan, D.,
  Doll{\'a}r, P., Zitnick, C.L.: Microsoft coco: Common objects in context. In:
  European conference on computer vision. pp. 740--755. Springer (2014)

\bibitem{liu2016ssd}
Liu, W., Anguelov, D., Erhan, D., Szegedy, C., Reed, S., Fu, C.Y., Berg, A.C.:
  Ssd: Single shot multibox detector. In: European conference on computer
  vision. pp. 21--37. Springer (2016)

\bibitem{long2015fully}
Long, J., Shelhamer, E., Darrell, T.: Fully convolutional networks for semantic
  segmentation. In: Proceedings of the IEEE conference on computer vision and
  pattern recognition. pp. 3431--3440 (2015)

\bibitem{nedic2001incremental}
Nedic, A., Bertsekas, D.P.: Incremental subgradient methods for
  nondifferentiable optimization. SIAM Journal on Optimization  \textbf{12}(1),
   109--138 (2001)

\bibitem{nowozin2014optimal}
Nowozin, S.: Optimal decisions from probabilistic models: the
  intersection-over-union case. In: Proceedings of the IEEE Conference on
  Computer Vision and Pattern Recognition. pp. 548--555 (2014)

\bibitem{parkhi12a}
Parkhi, O.M., Vedaldi, A., Zisserman, A., Jawahar, C.V.: Cats and dogs. In:
  IEEE Conference on Computer Vision and Pattern Recognition (2012)

\bibitem{rahman2016optimizing}
Rahman, M.A., Wang, Y.: Optimizing intersection-over-union in deep neural
  networks for image segmentation. In: International symposium on visual
  computing. pp. 234--244. Springer (2016)

\bibitem{redmon2017yolo9000}
Redmon, J., Farhadi, A.: Yolo9000: better, faster, stronger. In: Proceedings of
  the IEEE conference on computer vision and pattern recognition. pp.
  7263--7271 (2017)

\bibitem{ren2015faster}
Ren, S., He, K., Girshick, R., Sun, J.: Faster r-cnn: Towards real-time object
  detection with region proposal networks. In: Advances in neural information
  processing systems. pp. 91--99 (2015)

\bibitem{rosasco2004loss}
Rosasco, L., Vito, E.D., Caponnetto, A., Piana, M., Verri, A.: Are loss
  functions all the same? Neural Computation  \textbf{16}(5),  1063--1076
  (2004)

\bibitem{russakovsky2015imagenet}
Russakovsky, O., Deng, J., Su, H., Krause, J., Satheesh, S., Ma, S., Huang, Z.,
  Karpathy, A., Khosla, A., Bernstein, M., et~al.: Imagenet large scale visual
  recognition challenge. International journal of computer vision
  \textbf{115}(3),  211--252 (2015)

\bibitem{schaul2013no}
Schaul, T., Zhang, S., LeCun, Y.: No more pesky learning rates. In:
  International Conference on Machine Learning. pp. 343--351 (2013)

\bibitem{simonyan2014very}
Simonyan, K., Zisserman, A.: Very deep convolutional networks for large-scale
  image recognition. arXiv preprint arXiv:1409.1556  (2014)

\bibitem{sutskever2012imagenet}
Sutskever, I., Hinton, G.E., Krizhevsky, A.: Imagenet classification with deep
  convolutional neural networks. Advances in neural information processing
  systems pp. 1097--1105 (2012)

\bibitem{szegedy2014scalable}
Szegedy, C., Reed, S., Erhan, D., Anguelov, D., Ioffe, S.: Scalable,
  high-quality object detection. arXiv preprint arXiv:1412.1441  (2014)

\bibitem{tieleman2014rmsprop}
Tieleman, T., Hinton, G.: Rmsprop gradient optimization. URL http://www. cs.
  toronto. edu/tijmen/csc321/slides/lecture\_slides\_lec6. pdf  (2014)

\bibitem{udacity}
Udacity: Self driving car.
  \url{https://github.com/udacity/self-driving-car/tree/master/annotations}
  (2017)

\bibitem{pmlr-v70-xu17a}
Xu, Y., Lin, Q., Yang, T.: Stochastic convex optimization: Faster local growth
  implies faster global convergence. In: Precup, D., Teh, Y.W. (eds.)
  Proceedings of the 34th International Conference on Machine Learning.
  Proceedings of Machine Learning Research, vol.~70, pp. 3821--3830. PMLR,
  International Convention Centre, Sydney, Australia (06--11 Aug 2017),
  \url{http://proceedings.mlr.press/v70/xu17a.html}

\bibitem{DBLP:journals/corr/YuJWCH16}
Yu, J., Jiang, Y., Wang, Z., Cao, Z., Huang, T.S.: Unitbox: An advanced object
  detection network. CoRR  \textbf{abs/1608.01471} (2016),
  \url{http://arxiv.org/abs/1608.01471}

\end{thebibliography}
\end{document}